%% file: main.tex
\documentclass[10pt]{article} 
\usepackage[preprint]{tmlr}

\input{math_commands.tex}

\usepackage{float}  
\usepackage{hyperref}
\hypersetup{pdftitle={Rank Measures of Fine-Tuning Updates Need a Noise Baseline},pdfauthor={Muhammad Usman Altaf}}
\usepackage{url}
\usepackage{graphicx}
\usepackage{booktabs}
\usepackage{amssymb}

\newcommand{\dW}{\Delta W}
\newcommand{\thresholdrank}{\rho_{0.01}}
\newcommand{\energyrank}{k_{0.9}}
\newcommand{\entropyrank}{\mathrm{eRank}}
\newcommand{\stablerank}{\mathrm{sr}}

\title{Rank Measures of Fine-Tuning Updates\\Need a Noise Baseline}

\author{\name Muhammad Usman Altaf \email mu3931368@gmail.com \\
      \addr Independent Researcher}

\def\month{MM}  
\def\year{YYYY} 
\def\openreview{\url{https://openreview.net/forum?id=XXXX}} 

\begin{document}

\maketitle

\input{sections/abstract}
\input{sections/introduction}
\input{sections/background}
\input{sections/setup}
\input{sections/results}
\input{sections/timber}
\input{sections/related}
\input{sections/discussion}

\bibliography{main}
\bibliographystyle{tmlr}

\appendix
\section*{Appendix}
\input{sections/appendix}

\end{document}

%% file: math_commands.tex
\usepackage{amsmath,amsfonts,bm}

\def\eqref#1{equation~\ref{#1}}

\def\1{\bm{1}}

\DeclareMathAlphabet{\mathsfit}{\encodingdefault}{\sfdefault}{m}{sl}
\SetMathAlphabet{\mathsfit}{bold}{\encodingdefault}{\sfdefault}{bx}{n}

\newcommand{\R}{\mathbb{R}}



%% file: sections/abstract.tex
\begin{abstract}
Studies of what fine-tuning does to a language model often summarise the
weight update $\dW$ by its rank. We compare the rank measures used for this
against the same measures on a random matrix of the same shape, and against
two harder references: the pretrained weights, and matched fine-tunes
without alignment content. Measures that count directions put our
fine-tuning updates only a little below noise; stable rank, computed from
the same singular values, puts them far below. Both are right: an update has
a few strong directions on top of a long tail. The pattern holds for thirteen
released pairs up to 8B parameters, and it is not a mark of alignment: with
training held fixed, instruction data does not stand out from Wikipedia or
code, DPO gives the same shape whether its preference labels are real or
half swapped at random (one model, modest preference strength), and
scrambled text gives the most concentrated update of all. Applied to a recent
claim that post-training is superficial because the effective rank of the
weights barely changes, the baselines show that a random update of the same
size leaves that rank almost unchanged while wrecking the model, and that an
edit built on the effective rank of the update has no measurable effect in
our tests. Rank measures of weight updates need a noise baseline, and claims
about alignment need a control without alignment content.
\end{abstract}

%% file: sections/introduction.tex
\section{Introduction}
\label{sec:intro}

When a language model is fine-tuned, each weight matrix $W$ changes by an
update $\dW$.\footnote{Large language model tools were used to assist with
editing the prose and with writing analysis code; all ideas, experiments,
analyses and claims are the author's own.} A natural way to ask what fine-tuning did is to ask how many
independent directions $\dW$ uses, that is, its rank. A low-rank update
suggests a small, targeted change; a full-rank one suggests a diffuse change.

The question matters most for alignment. The superficial alignment hypothesis
\citep{zhou2023lima} holds that alignment mostly teaches a model which style
to answer in, and output-level evidence supports it \citep{lin2024urial}. If
that is true, one might expect alignment to leave a small, structured
footprint in the weights, and recent work has looked for one in the rank of
$\dW$ and of $W$ \citep{wu2025timber}. Related work reports that fine-tuning
updates are ``high rank'' \citep{biderman2024lora} or ``nearly full-rank''
\citep{mukherjee2025rl}.

Real updates are full rank in the strict sense, so these readings rest on
softer measures: the numerical rank, the number of directions that hold 90\%
of the energy \citep{biderman2024lora}, or the entropy effective rank
\citep{roy2007effective, wu2025timber}. \citet{biderman2024lora} also compared
against noise: the spectrum of a fully fine-tuned weight matrix looks like that
of the pretrained one plus low-magnitude Gaussian noise, and they concluded
that nothing about it is unusual. We began this project with the same
impression. Counting the singular values of a supervised fine-tuning (SFT)
update above 1\% of the
largest gave 97\% of full rank, and the difference between two random
initialisations of the same model, which is pure noise, gave 99.6\%. By that
measure, fine-tuning looks almost like noise.

It is not. Stable rank \citep{rudelson2007sampling}, computed from the same
singular values, puts the update 4.5 times below noise, and its top direction
holds 6.6 times as much of the energy (4.2\% against 0.6\%). The two answers
differ because the
measures weigh different parts of the spectrum, and because measures of $W$
cannot see much of $\dW$ at all. Comparing against noise is an old idea in
statistics \citep{horn1965parallel, gavish2014optimal}, has been used to
study trained weights \citep{martin2021implicit, thamm2022random}, and has
been applied to fine-tuning updates before: \citet{ryu2023efficient} plot the
spectrum of an update against a Marchenko--Pastur reference, and
\citet{biderman2024lora} against Gaussian noise. What has not been done is to
apply it measure by measure, which is where the disagreement is, with matched
controls for what the update was trained on. That is what this paper does.

Our findings:

\begin{enumerate}
  \item \textbf{Counting measures compress the gap to noise.} On our SFT,
  Wikipedia, code and direct preference optimisation (DPO) fine-tunes, the
  threshold count is only 2--5\%
  below a random matrix of the same shape, and eRank and the 90\% energy rank
  are 12--35\% below it; stable rank is 78--94\% below. Along the whole family
  of R\'enyi ranks the gap grows with how much weight a measure gives to the
  top directions, then levels off, as it must for any spectrum with a heavier
  head than noise; the point is how large the disagreement is
  (\S\ref{sec:metrics}).
  \item \textbf{Updates have a few strong directions and a long tail}, and on
  average are more concentrated than the pretrained weights they are added to
  (\S\ref{sec:headtail}).
  \item \textbf{The shape does not single out alignment training.} With the
  amount of training held fixed, the data does shape an update (code is the
  least concentrated, the more so with more training), but instruction SFT
  does not stand out from Wikipedia or Python code. DPO gives the same shape
  whether it learns real preferences or randomly swapped ones that teach
  nothing; a signed comparison of the updates does tell the two apart. Text with its word order scrambled gives
  the most concentrated update of all (\S\ref{sec:controls}).
  \item \textbf{It holds for released models}, thirteen base/instruct,
  post-training-stage and base/specialist pairs up to 8B parameters. On three
  bf16 updates (Qwen2.5-1.5B and the OLMo 2 DPO and RL stages), much of what
  the threshold count counts may be rounding (\S\ref{sec:released}).
  \item \textbf{It changes how earlier evidence reads.} An unchanged effective
  rank of $W$ after post-training, read by \citet{wu2025timber} as a sign that
  post-training is superficial, is what a random update of the same size also
  produces, even one that cuts the model's GSM8K accuracy from 60\% to 5\%.
  The edit they build on the effective rank of the update has no measurable
  effect on GSM8K, nor, within about three points, on their own benchmark,
  GPQA-Diamond (\S\ref{sec:timber}).
\end{enumerate}

The practical message is short: report rank measures of weight updates next to
the same measure on noise, check that low-precision training or storage did
not erase the update, and test claims about alignment against fine-tuning
without alignment content.

%% file: sections/background.tex
\section{Measuring the rank of an update}
\label{sec:background}

Let $\dW \in \R^{m\times n}$ be the update to one weight matrix,
$r = \min(m,n)$, and $\sigma_1 \geq \dots \geq \sigma_r$ its singular values.
Write $p_i = \sigma_i^2 / \sum_j \sigma_j^2$ for the share of the update's
energy (squared Frobenius norm) in direction $i$. We report every measure as a
fraction of $r$, so that matrices of different shapes can be averaged.

\paragraph{Five measures.} In plain words, each asks ``how many directions
does this update really use?'', and they differ in how much the strongest
directions count.
\begin{itemize}
  \item \textbf{Threshold count} $\thresholdrank$: how many $\sigma_i$ are at
  least $0.01\,\sigma_1$. We use it as the simplest example of counting; the
  numerical rank reported by \citet{mukherjee2025rl} counts in the same way,
  with a tolerance they do not state (by default it is set by floating-point
  precision).
  \item \textbf{Energy rank} $\energyrank$: the smallest $k$ with
  $\sum_{i\leq k} p_i \geq 0.9$, the measure of \citet{biderman2024lora}.
  \item \textbf{Entropy effective rank} $\entropyrank$
  \citep{roy2007effective}: $\exp\bigl(-\sum_i q_i \log q_i\bigr)$ with
  $q_i = \sigma_i/\sum_j \sigma_j$, as used by \citet{wu2025timber}.
  \item \textbf{Participation ratio}: $1/\sum_i p_i^2$.
  \item \textbf{Stable rank} $\stablerank$ \citep{rudelson2007sampling}:
  $\|\dW\|_F^2/\|\dW\|_2^2 = 1/p_1$.
\end{itemize}
All five ignore the overall scale of $\dW$ (they are scale-invariant), which matters because the updates
we compare differ in size by orders of magnitude.

\paragraph{How the measures relate.} Three of the five are effective ranks of
one kind: the exponential of a R\'enyi entropy of $p$ of order $\alpha$,
$R_\alpha = \bigl(\sum_i p_i^\alpha\bigr)^{1/(1-\alpha)}$. Order 0 counts the
nonzero directions (the threshold count is a thresholded version of it),
order 2 is the participation ratio, and order $\infty$ is the stable rank.
The higher the order, the more the top directions dominate. This is why the
measures can disagree about the same matrix: a spectrum with a few strong
directions and a long tail of weak ones has many directions to count but most
of its weight on top. The other two are close relatives rather than members:
eRank uses the Shannon entropy of $q \propto \sigma$ instead of
$p \propto \sigma^2$, and the energy rank counts directions up to a fixed share
of $p$. We list them between the threshold count and the participation ratio,
which is where their values fall for our updates.

Note that $R_\alpha$ can only fall as $\alpha$ grows, for any matrix, so the
order of the raw values tells us nothing. What does carry information is how
far each measure puts an update below noise.

\paragraph{The gap to noise.} For each measure we report
$1 - \bar{x}_{\text{update}}/\bar{x}_{\text{noise}}$, where $\bar{x}$ is the
mean of the measure over all matrices of a model, whatever their shape: 0
means the update looks like noise, and values near 1 mean it is far more
concentrated. (The noise value itself depends on the shape, from 0.25 to 0.53
for stable rank, Table~\ref{tab:mp}; Table~\ref{tab:pertype} gives the values
by matrix type.) Intervals come from a bootstrap over layers
(\S\ref{sec:setup}), and ratios quoted in the text, such as ``4.5 times below
noise'', use the unrounded means. The measures have different ceilings (the threshold count of
noise is already 0.996), so the gap is best compared across updates for one
measure, and only roughly across measures.

\paragraph{The noise baseline.} A value of any of these measures only means
something next to its value for a matrix with no structure. For our own runs
the baseline is the difference between two independent random initialisations
of the same model; for released models, one Gaussian draw for each matrix,
with the same shape. The spectrum of such a matrix follows the
Marchenko--Pastur law, whose upper edge (the Bai--Yin limit,
\citealp{bai2010spectral}) gives a stable rank close to
\[
  \frac{mn}{(\sqrt{m}+\sqrt{n})^2},
\]
a quarter of $r$ for square matrices. Our
random-init pairs match this to within 0.01 for every matrix shape of
Qwen2.5-0.5B (Table~\ref{tab:mp} in Appendix~\ref{app:sensitivity}), and the
two baselines agree: on Qwen2.5-0.5B the stable rank is 0.436 for the
random-init pair and 0.434 for the Gaussian matrices.

Beating this baseline is necessary but not sufficient. A Gaussian matrix is
the least structured baseline possible, and any difference between two
trained matrices beats it: the pretrained weights $W$ themselves have a
stable rank well below noise, and so does the difference between two
\emph{unrelated} trained checkpoints (\S\ref{sec:setup}). The noise value
tells the reader how to read a number; it does not by itself show that an
update is special. So we use two further references: the pretrained weights
$W$, and fine-tunes on data with no alignment content, trained the same way
(\S\ref{sec:controls}).

\paragraph{The rounding floor.} Released checkpoints are usually stored in
bf16, which rounds every weight to 8 significant bits. The rounding adds a
small noise matrix to $\dW$. For each released pair we simulate it (uniform
error up to half a bf16 unit at each weight, independently for both
checkpoints). By Weyl's inequality, rounding cannot create a direction of
$\dW$ stronger than its own largest singular value, so we report the share of
the update's singular values above that ``rounding floor''. The simulation is
a generous model of rounding: when a weight barely moves, its two rounding
errors are not independent, and the stored difference is exactly zero. So we
also report the direct evidence, the share of entries of $\dW$ that are
exactly zero, meaning that the two checkpoints store the identical bf16
value.

\paragraph{Measures of $W$ rather than $\dW$.} Some studies measure the full
weight matrix before and after fine-tuning. By Weyl's inequality
\citep{horn2012matrix}, each singular value of $W + \dW$ is within
$\|\dW\|_2$ of the corresponding one of $W$. When the update is small
compared with $W$, the large singular values of $W$ barely move; the small
ones can move more relative to their size, so the bound alone does not settle
how much a measure of $W$ can change. We therefore test it directly against a
random update of the same size, $W + G\,\|\dW\|_F/\|G\|_F$ with $G$
Gaussian.

%% file: sections/setup.tex
\section{Setup}
\label{sec:setup}

\paragraph{Our fine-tunes.} We fine-tune Qwen2.5-0.5B base \citep{qwen2.5}
and SmolLM2-360M base \citep{allal2025smollm2}, all parameters, four ways
each, with the same optimizer settings and the same number of training tokens
(about 1.75M for Qwen and 1.9M for SmolLM2 per epoch, counted after
truncation, for three epochs):
\begin{itemize}
  \item \textbf{SFT} on 10k instruction examples from Alpaca-cleaned, a
  community-cleaned version of Alpaca \citep{taori2023alpaca}, with the loss
  on answer tokens only;
  \item \textbf{wikitext}: plain language-model training on wikitext-103
  \citep{merity2017pointer}, a corpus of Wikipedia articles;
  \item \textbf{code}: the same on deduplicated Python files from CodeParrot
  \citep{codeparrot};
  \item \textbf{shuffled words}: the same wikitext documents with the word order
  scrambled inside each document, so the words and their frequencies are
  unchanged but there is no structure left to learn.
\end{itemize}
SFT, wikitext and code were each run twice on both models, and DPO
(\S\ref{sec:controls}) twice on Qwen, with a different seed and a different
data sample each time.
Matching tokens does not match optimizer steps: Python files are long, so the
code runs take about 700 steps against about 1,850 for SFT and 1,900--2,100
for wikitext and shuffled words. A further code run on each model matches the
steps of SFT instead, at the cost of 2.5--2.7 times more code tokens.
Appendix~\ref{app:training} lists every setting (Table~\ref{tab:hparams}) and
the steps, loss and update size of every run (Table~\ref{tab:runs}).

\paragraph{Precision.} All runs fit on one 6\,GB GPU. The weights are kept in
fp32 and only the compute runs in bf16, with 8-bit AdamW
\citep{dettmers2022optimizers} and gradient checkpointing
\citep{chen2016checkpointing}. This matters: if the weights themselves are
stored in bf16, any update smaller than half a bf16 unit is rounded away
\citep{zamirai2020bfloat16, zhu2025path}. In a first round of our own runs that
did exactly this, 77--98\% of the weights never changed, and the measured
``update'' was an artefact of the rounding. Every run reported here is
checked after training: at most $1.01\times10^{-5}$ of the weights are unchanged. We
recommend this check to anyone measuring fine-tuning updates.

\paragraph{Released models.} We measure thirteen official pairs with no
training, in three groups. (i) Base/instruct releases: SmolLM2 135M, 360M and
1.7B; Qwen2.5 0.5B and 1.5B; Qwen3 0.6B and 8B \citep{yang2025qwen3}, the two
Qwen3 pairs being among those studied by \citet{wu2025timber}. (ii) The three
post-training stages of OLMo 2 1B \citep{olmo2024olmo2}, base to SFT, SFT to
DPO and DPO to reinforcement learning (RL), whose training data is public.
(iii) Large updates without
alignment content: Qwen2.5-Math-1.5B \citep{yang2024qwenmath} and
Qwen2.5-Coder 0.5B and 1.5B \citep{hui2024qwencoder} against their bases.

Before measuring a pair we check that the second checkpoint descends from the
first: the cosine between the two sets of weights is 0.99 or more for all
seven base/instruct pairs and the OLMo stages, and 0.20--0.60 for the
specialists, which were trained on far more data. The check matters. The
released OLMo 2 1B SFT checkpoint (revision \texttt{0d85a3d}) has cosine
0.00 with the main revision of the released base model (\texttt{a1847df}),
over all 112 matrices: the two do not even share an initialisation. The base
model's mid-training checkpoints are published as branches. The final
checkpoint of the third mid-training run
(\texttt{stage2-ingredient3-step23852-tokens51B}, revision \texttt{f83cd64})
has cosine 0.995 with the SFT model; the first two runs, which the model card
describes as exploratory, reach 0.991. For the 1B model only the third run was used, with
no merging, so we take its final checkpoint as the base of the SFT stage. The
difference between the unrelated pair still looks like a fine-tuning update:
its stable rank is 0.077 against 0.336 for noise, 4.4 times below, close to
the 0.080 of the real SFT update. The model card does not explain the
difference, and we have not confirmed its cause with the maintainers; we
report what the public checkpoints contain.

\paragraph{What we measure.} We measure every attention and MLP projection
(\texttt{q}, \texttt{k}, \texttt{v}, \texttt{o}, \texttt{gate}, \texttt{up},
\texttt{down}): 168 matrices per Qwen2.5-0.5B model, up to 252 for Qwen3-8B.
Embeddings and the output head are excluded. We report means over matrices.
Matrices in the same layer are not independent, so 95\% intervals come from a
bootstrap that resamples whole layers, and comparisons between runs on the
same model are paired matrix by matrix. These intervals describe the spread
over layers within one run, not the variation between training runs; for
that we have the second seeds.

%% file: sections/results.tex
\section{What the rank measures say}
\label{sec:results}

\subsection{Counting measures compress the gap to noise}
\label{sec:metrics}

\begin{table}[t]
\caption{Rank measures as a fraction of the maximum rank, and the share of
energy in the top direction, averaged over all attention and MLP matrices.
``Noise'' is the random-init pair; ``pretrained $W$'' is the base model's own
weights. The DPO updates are measured from the SFT model they start from.}
\label{tab:measures}
\begin{center}
\small
\begin{tabular}{llcccccc}
\toprule
model & matrix & threshold & eRank & 90\% energy & particip. & stable & top-1 energy \\
\midrule
Qwen2.5-0.5B & SFT update            & 0.973 & 0.816 & 0.541 & 0.381 & 0.097 & 4.2\% \\
             & wikitext update       & 0.972 & 0.810 & 0.531 & 0.365 & 0.093 & 4.0\% \\
             & code update           & 0.975 & 0.817 & 0.543 & 0.374 & 0.094 & 3.8\% \\
             & shuffled-words update & 0.941 & 0.745 & 0.440 & 0.202 & 0.042 & 9.0\% \\
             & DPO update            & 0.966 & 0.810 & 0.536 & 0.295 & 0.053 & 5.8\% \\
             & swapped-DPO update    & 0.967 & 0.811 & 0.537 & 0.304 & 0.055 & 5.7\% \\
             & noise                 & 0.996 & 0.928 & 0.716 & 0.754 & 0.436 & 0.6\% \\
             & pretrained $W$        & 0.976 & 0.846 & 0.590 & 0.537 & 0.178 & 2.2\% \\
\midrule
SmolLM2-360M & SFT update            & 0.951 & 0.755 & 0.439 & 0.145 & 0.025 & 8.3\% \\
             & wikitext update       & 0.948 & 0.744 & 0.425 & 0.111 & 0.020 & 9.6\% \\
             & code update           & 0.963 & 0.751 & 0.432 & 0.183 & 0.034 & 5.5\% \\
             & shuffled-words update & 0.863 & 0.629 & 0.275 & 0.040 & 0.009 & 21.0\% \\
             & noise                 & 0.996 & 0.908 & 0.659 & 0.668 & 0.354 & 0.5\% \\
             & pretrained $W$        & 0.982 & 0.841 & 0.562 & 0.488 & 0.140 & 1.6\% \\
\bottomrule
\end{tabular}
\end{center}
\end{table}

\begin{figure}[t]
\begin{center}
\includegraphics[width=\linewidth]{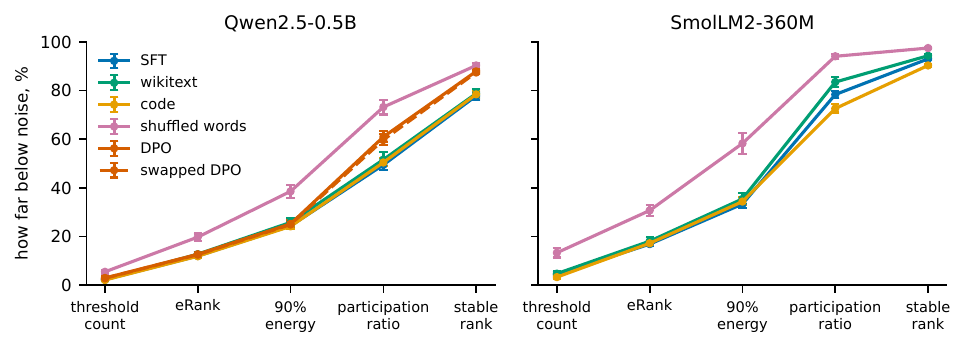}
\end{center}
\caption{How far below noise each measure puts each update,
$1-\text{update}/\text{noise}$, with 95\% layer-bootstrap intervals. Measures
are ordered by how much weight they give to the top directions. The gap grows
from a few percent for the threshold count to 78--94\% for stable rank, and
to 90--97\% for the shuffled-words updates.}
\label{fig:gaps}
\end{figure}

Table~\ref{tab:measures} gives the five measures for our updates. By the
threshold count, the Qwen SFT, wikitext and code updates are at 97\% of full
rank, close to the 99.6\% of noise. Stable rank tells a different story: every
update is several times below noise, 4.5 times for the Qwen SFT update, 10 to
17 times for the SmolLM2 updates, and up to 39 times for shuffled words.

Figure~\ref{fig:gaps} shows how the measures fit together. Ordered from the
threshold count to stable rank, each measure puts the updates further below
noise: about 2\% for the threshold count, 12\% for eRank, 25\% for the 90\%
energy rank, 50\% for the participation ratio and 78\% for stable rank on
Qwen, and the same order on SmolLM2. The values for noise fall along the same
order, and what grows is the distance between update and noise. Along the
whole family of R\'enyi ranks the gap grows steadily with the order, then
levels off (Figure~\ref{fig:renyi} in Appendix~\ref{app:sensitivity}). This is
expected for any spectrum with a heavier head than noise, so we read it less
as a finding than as a warning: the answer depends on which order one
happens to pick. Every interval excludes zero, so all five measures do detect
that the updates are not random. The problem is the size of the gap. The two
measures used in the literature, eRank and the energy rank, put the SFT,
wikitext, code and DPO updates 12--18\% and 24--35\% below noise; stable rank
puts them 78--94\% below. The threshold count is also arbitrary: at
$0.1\,\sigma_1$ instead of $0.01\,\sigma_1$ the Qwen SFT update sits 18\%
below noise instead of 2\% (Table~\ref{tab:threshold}). A reader who sees
``97\% of full rank'' without the noise value next to it will conclude that the
update is diffuse, which is what we concluded at first. The same holds for
every matrix type (Table~\ref{tab:pertype}).

\subsection{A few strong directions and a long tail}
\label{sec:headtail}

\begin{figure}[t]
\begin{center}
\includegraphics[width=\linewidth]{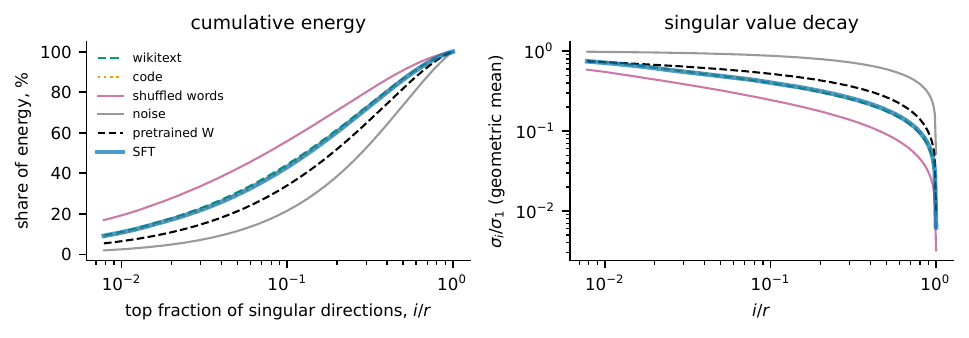}
\end{center}
\caption{Where the energy sits, averaged over the 168 matrices of
Qwen2.5-0.5B. Left: share of energy in the top fraction of directions. Right:
singular values relative to the largest. Updates put much more energy in their
top directions than noise does, then decay slowly over a long tail. The
pretrained weights (dashed) sit between the updates and noise.}
\label{fig:spectrum}
\end{figure}

Figure~\ref{fig:spectrum} shows why the measures disagree. At the top of the
spectrum the updates are far more concentrated than noise: the top direction of
the Qwen SFT update holds 4.2\% of its energy against 0.6\% for noise. Below
that the spectrum decays slowly, and this long tail still holds most of the
energy. Stable rank is dominated by the top, so it sees the concentration; the
counting measures count directions, most of which are in the tail, so they see
a nearly full-rank matrix. Both descriptions are true of the same matrix:
fine-tuning updates are high rank \citep{biderman2024lora} and they are
concentrated. Long-tailed update spectra have been noted before, in work on
compressing updates \citep{ping2024deltacome}; what the noise baseline adds is
how far from random the head of the spectrum is. Nor is the result driven by
one outlying direction: without the top direction, stable rank is 0.126 for
the Qwen SFT update against 0.440 for noise (Table~\ref{tab:runs}; the noise
value is in Appendix~\ref{app:training}).

The pretrained weights are a harder reference than noise, since they are
themselves far from random, and the updates are more concentrated than them
too, on average. The stable rank of the pretrained weights is 0.178 on Qwen
and 0.140 on SmolLM2, against 0.093--0.097 and 0.020--0.034 for the SFT,
wikitext and code updates: 46--48\% below the weights on Qwen and 75--86\% on
SmolLM2, with layer-bootstrap intervals well clear of zero. This does not hold
for every matrix type: on Qwen, for the \texttt{down} projection, the SFT update and the
weights are level (Table~\ref{tab:pertype}). The threshold count cannot see
the difference either: it puts the Qwen updates and the pretrained weights at
the same 97--98\%.

\subsection{Instruction and preference training leave no special mark}
\label{sec:controls}

\begin{table}[t]
\caption{Differences in stable rank between runs on the same model, paired
matrix by matrix, with 95\% layer-bootstrap intervals. The intervals describe
the spread over layers within one run, not between training runs; ``second
seed'' repeats a run with a different seed and data sample. ``Steps matched''
trains on code for as many optimizer steps as SFT, which is 2.5--2.7 times more code
tokens.}
\label{tab:paired}
\begin{center}
\small
\begin{tabular}{lcc}
\toprule
difference & Qwen2.5-0.5B & SmolLM2-360M \\
\midrule
SFT $-$ wikitext            & $+0.004$ ($-0.001$, $+0.010$) & $+0.005$ ($+0.003$, $+0.006$) \\
SFT $-$ code                & $+0.003$ ($-0.000$, $+0.006$) & $-0.009$ ($-0.011$, $-0.008$) \\
code $-$ wikitext           & $+0.002$ ($-0.004$, $+0.007$) & $+0.014$ ($+0.012$, $+0.016$) \\
shuffled words $-$ SFT      & $-0.054$ ($-0.062$, $-0.047$) & $-0.016$ ($-0.018$, $-0.014$) \\
code, steps matched $-$ code & $+0.020$ ($+0.018$, $+0.023$) & $+0.007$ ($+0.007$, $+0.008$) \\
swapped DPO $-$ DPO         & $+0.002$ ($+0.000$, $+0.004$) & -- \\
\midrule
second seed $-$ first, SFT      & $-0.001$ ($-0.005$, $+0.001$) & $-0.000$ ($-0.001$, $-0.000$) \\
second seed $-$ first, wikitext & $-0.001$ ($-0.005$, $+0.003$) & $+0.000$ ($-0.000$, $+0.000$) \\
second seed $-$ first, code     & $-0.000$ ($-0.001$, $+0.001$) & $-0.000$ ($-0.001$, $+0.000$) \\
second seed $-$ first, DPO      & $+0.001$ ($-0.001$, $+0.002$) & -- \\
\bottomrule
\end{tabular}
\end{center}
\end{table}

A concentrated SFT update says nothing about instruction tuning unless
fine-tuning on other data looks different. With the model, optimizer and
number of training tokens held fixed, SFT does not stand out. On Qwen, SFT,
wikitext and code give stable ranks of 0.097, 0.093 and 0.094; every paired
difference between them is under 0.005 and its interval includes zero
(Table~\ref{tab:paired}), and a second seed moves each by at most 0.0014. On
SmolLM2 the data does leave a mark: code is the least concentrated (0.034),
wikitext the most (0.020) and SFT sits between them (0.025). These
differences are real, not seed noise: a second seed of each SmolLM2 run moves
its stable rank by at most 0.0003. Nor are they an artefact of the code runs
taking 2.6--2.7 times fewer optimizer steps than SFT (\S\ref{sec:setup}): training on
code for as many steps as SFT, which is 2.5--2.7 times more code tokens, moves the
code update further from SFT, not closer, on both models (stable rank 0.114
on Qwen, 0.042 on SmolLM2). So what the data teaches, and how much of it, does
shape the update. What it does not do is single out instruction tuning: in
both models SFT does not stand out from the ordinary language-model runs.

Similar stable rank means a similar spectral shape, not the same update, so we
also compared the directions. For each pair of runs we measured the overlap of
the top-16 left singular subspaces of each matrix's update, averaged over
matrices (1 would be identical subspaces; two random subspaces give 0.02--0.04;
Appendix~\ref{app:overlap}). Two runs on the same kind of data with different
seeds overlap by 0.26--0.32 on Qwen and 0.34--0.51 on SmolLM2. Runs on
different data overlap less, 0.20--0.22 on Qwen and 0.12--0.13 on SmolLM2, but
still four to six times more than random. So the data steers which directions
an update uses, more than it changes the update's shape.

Preference training gives the same answer. Starting from the Qwen SFT model,
we ran DPO \citep{rafailov2023dpo} on 10k UltraFeedback pairs
\citep{cui2024ultrafeedback}, and a control on the same pairs with chosen and
rejected swapped in a random half of them, in the spirit of random-label
controls \citep{zhang2017rethinking}. The prompts, responses and amount of
training are identical, but there is no net preference to learn; the control
does share the unswapped half of its pairs with the DPO run, so it removes the
preference signal, not everything the two runs see in common
(Appendix~\ref{app:training}). On about 190 held-out pairs the DPO model
prefers the chosen answer 65\% of the time (74\% for a second seed); the
control is at chance, 51\%. Yet the two updates are alike on every measure
(Table~\ref{tab:measures}): stable rank 0.053 for DPO and 0.055 for the
control, against 0.436 for noise, and the two updates have the same size
(Table~\ref{tab:runs}). The paired difference,
$+0.002$, is larger than the gap between the two DPO seeds ($+0.001$) but
under 4\% of the value and about $1/200$ of the distance to noise
(Table~\ref{tab:paired}). Like the SFT update, the DPO update is more
concentrated than the pretrained weights and far more than noise, and at the
strength of preference learning we reach, almost none of that shape comes from
having learned the preferences.

The directions are another matter. The subspace overlap cannot separate these
runs: the real DPO update overlaps with the swapped control (0.27) almost
exactly as much as with a second seed of itself (0.28). But that overlap is the
same for an update and its negative, and swapping a pair reverses its preference
signal. The plain cosine between updates keeps the sign
(Appendix~\ref{app:overlap}). It is 0.084 (95\% interval 0.082, 0.087) between
the two real DPO seeds, 0.040 (0.039, 0.042) between the real run and the swapped
control, and 0.001 ($-$0.000, 0.002) between the second seed and the swapped
control. So the real runs agree more with each other than with the control. We
do not read this as a measurement of the preference signal alone: the two real
runs share 16\% of their training pairs, the swapped run uses all the pairs of
the first, and there is one swapped run on one model.

The surprise is the shuffled-words control. Text with its word order
scrambled, which has little to learn beyond word frequencies, gives the most
concentrated update on both models: stable rank 0.042 on Qwen and 0.009 on
SmolLM2, with 9\% and 21\% of the energy in a single direction. This holds in
every quarter of the network (Table~\ref{tab:shuffled}). The model fits this
data poorly (training loss 5.4, against 2.2 for wikitext;
Table~\ref{tab:runs}), and we do not know why that gives a more concentrated
update. One explanation we tested, that the update mainly pushes the output
towards frequent tokens, does not single it out: the top directions of the
SFT and code updates line up with token frequency in late layers too
(Appendix~\ref{app:shuffled}). We report it as an observation: the
concentration of a fine-tuning update depends on the training process and the
data, not on whether the data teaches instruction following or preferences.

\subsection{Released models}
\label{sec:released}

\begin{figure}[t]
\begin{center}
\includegraphics[width=\linewidth]{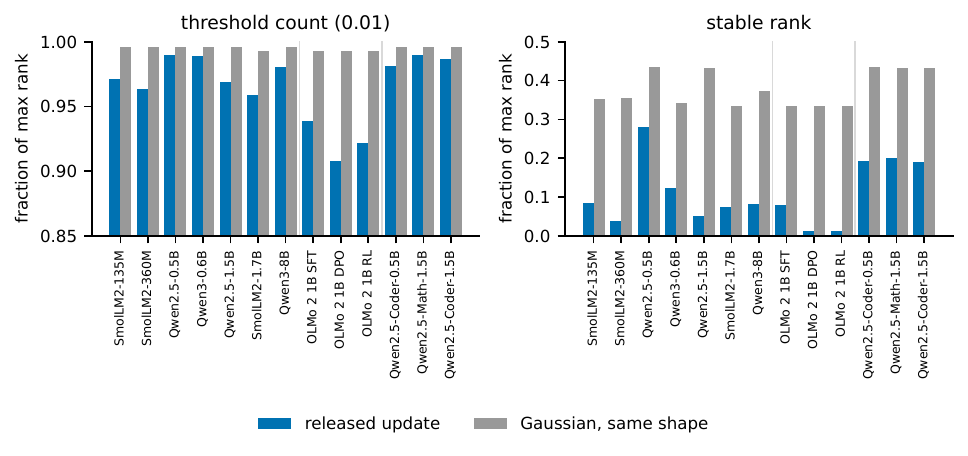}
\end{center}
\caption{Released updates (blue) against a Gaussian matrix of the same shape
(grey), in three groups: base/instruct releases, the OLMo 2 post-training
stages, and code/math specialists. Left: the threshold count (axis starts at
0.85). Right: stable rank.}
\label{fig:released}
\end{figure}

\begin{table}[t]
\caption{Released pairs, means over matrices. ``Above floor'' is the share of
the update's singular values larger than the largest singular value of the
simulated bf16 rounding error: directions that rounding alone cannot produce
(\S\ref{sec:background}). It is not computed (--) for OLMo 2 base to SFT,
whose base, a mid-training checkpoint, is stored in fp32. The last column drops the top direction, to
check that one outlier does not drive the result.}
\label{tab:released}
\begin{center}
\small
\begin{tabular}{lcccccc}
\toprule
update & threshold & noise & above floor & stable rank & noise & stable w/o top \\
\midrule
SmolLM2-135M instruct & 0.971 & 0.996 & 96\% & 0.086 & 0.354 & 0.119 \\
SmolLM2-360M instruct & 0.964 & 0.996 & 95\% & 0.039 & 0.356 & 0.059 \\
Qwen2.5-0.5B instruct & 0.990 & 0.996 & 97\% & 0.280 & 0.434 & 0.314 \\
Qwen3-0.6B            & 0.989 & 0.996 & 99\% & 0.124 & 0.342 & 0.145 \\
Qwen2.5-1.5B instruct & 0.969 & 0.996 & 78\% & 0.051 & 0.432 & 0.073 \\
SmolLM2-1.7B instruct & 0.959 & 0.993 & 91\% & 0.075 & 0.336 & 0.090 \\
Qwen3-8B              & 0.980 & 0.996 & 97\% & 0.082 & 0.373 & 0.098 \\
\midrule
OLMo 2 1B: base $\to$ SFT & 0.939 & 0.993 & --    & 0.080 & 0.336 & 0.100 \\
OLMo 2 1B: SFT $\to$ DPO  & 0.908 & 0.993 & 13\% & 0.012 & 0.336 & 0.023 \\
OLMo 2 1B: DPO $\to$ RL   & 0.922 & 0.993 & 3\% & 0.013 & 0.336 & 0.021 \\
\midrule
Qwen2.5-Coder-0.5B    & 0.981 & 0.996 & 100\% & 0.194 & 0.434 & 0.228 \\
Qwen2.5-Math-1.5B     & 0.990 & 0.996 & 100\% & 0.200 & 0.432 & 0.245 \\
Qwen2.5-Coder-1.5B    & 0.987 & 0.996 & 100\% & 0.191 & 0.432 & 0.234 \\
\bottomrule
\end{tabular}
\end{center}
\end{table}

Released instruct models go through much larger and more varied
post-training than our runs. The same pattern holds for all thirteen pairs
(Figure~\ref{fig:released}, Table~\ref{tab:released}). The threshold count
reads 91--99\%, close to noise; stable rank puts every update 1.6 to 27 times
below noise, and dropping the top direction changes little, so no single
outlier drives it. Every update also has tens to hundreds of directions above
the largest singular value a same-size Gaussian matrix would have
(Table~\ref{tab:released_more}).

For most pairs, bf16 rounding is not a concern: 91--100\% of their
directions are stronger than anything rounding alone can produce. For three
it is: 14\% of the entries of the Qwen2.5-1.5B update, and 39\% and 52\% of
the OLMo DPO and RL updates, are exactly zero, and only 78\%, 13\% and 3\% of
their directions rise above the simulated rounding floor
(Tables~\ref{tab:released} and~\ref{tab:released_more}). For these updates
much of what the threshold
count counts may be rounding. Rounding adds many tiny singular values, which
stable rank largely ignores.

The OLMo 2 stages show where post-training changes the weights. Almost all of
the change is in SFT: its update is 9.8\% of the size of the weights, has
stable rank 0.080 against 0.336 for noise, and the whole base-to-instruct
update has the same size and shape to within 0.001. The DPO and RL stages are
tiny, under half a percent of the weights, and only 13\% and 3\% of their
directions rise above the simulated rounding floor. Their very low stable
rank therefore reflects a few real directions on top of rounding, and we draw
no conclusion about preference training from them; our own DPO runs
(\S\ref{sec:controls}) are the controlled test.

The specialists give a mixed picture. Against an instruct release of the same
base, Qwen2.5-Coder-0.5B (stable rank 0.194) is more concentrated than
Qwen2.5-0.5B-Instruct (0.280), while Qwen2.5-Math-1.5B and
Qwen2.5-Coder-1.5B (0.200 and 0.191) are less concentrated than
Qwen2.5-1.5B-Instruct (0.051). The specialists differ from the instruct models
in both content and amount of training (their updates are about as large as
the weights themselves), so this is not a controlled comparison; our matched
runs in \S\ref{sec:controls} are. What the released models show is that
concentration varies with how much and what kind of training was done, and
that nothing in it sets instruct updates apart.

%% file: sections/timber.tex
\section{Case study: is post-training superficial?}
\label{sec:timber}

\citet{wu2025timber} compare released base and instruct models (Llama~3.1 and 3.2,
\citealp{grattafiori2024llama3}, and Qwen3) through the entropy effective rank
(eRank, \S\ref{sec:background}) of each weight matrix $W$ (not of $\dW$). They find it almost unchanged by post-training, at about 85\% of full
rank, and read this as weight-level evidence that post-training is
superficial. They also propose an edit, Timber: keep the top
$\lceil \entropyrank(\dW)\rceil$ singular directions of the update and scale
the rest by $\lambda<1$, with $\lambda=0.2$ recommended, which improves Mean@$k$
and pass@$k$ on their benchmarks. We check both with the baselines of
\S\ref{sec:background}.

\subsection{An unchanged effective rank is what any update of this size gives}

\begin{figure}[t]
\begin{center}
\includegraphics[width=\linewidth]{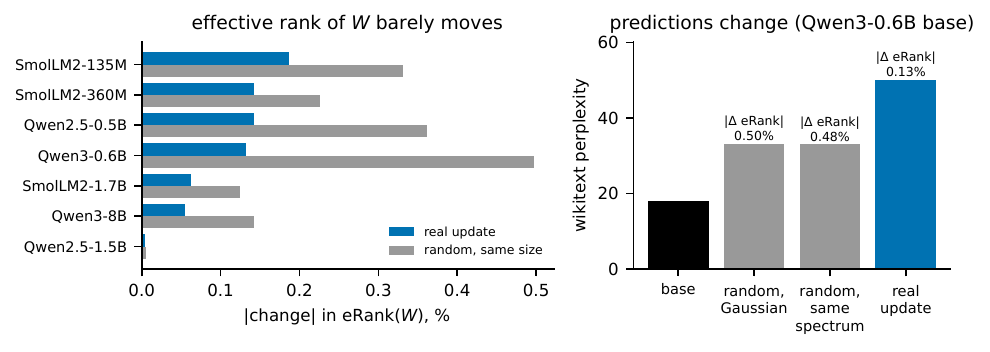}
\end{center}
\caption{Left: size of the change in the effective rank of the full weight
matrix $W$, $|\Delta\,\entropyrank(W)| / \entropyrank(W)$ averaged over
matrices, for the real base-to-instruct update (blue) and a random update of
the same Frobenius norm (grey). Right: perplexity on held-out
wikitext of Qwen3-0.6B base, unchanged, after a random update of the same size
as its real post-training update (Gaussian, or with the real update's singular
values), and after the real update, which gives the instruct model. The
numbers above the bars give the size of the change in eRank($W$) for each
update, as in the left panel.}
\label{fig:random}
\end{figure}

The measurement reproduces. On Qwen3-0.6B and Qwen3-8B, two of their pairs,
the effective rank of the base weights is 0.84 and 0.86 of full rank, and
post-training changes it by only 0.13\% and 0.06\%.

But a random update of the same size also leaves it almost unchanged
(Figure~\ref{fig:random}, left). On all seven base/instruct pairs, whose
updates are 1--13\% of the size of the weights, real and random updates both
move the effective rank of $W$ by at most 0.5\%. The real update moves it 1.6
to 3.8 times less than the random one, so post-training does change $W$ in a
structured way, but that is visible only next to the random comparison.

\begin{table}[t]
\caption{GSM8K accuracy, in percent (greedy, all 1{,}319 test problems), of Qwen3-0.6B with
a random update of the same size as its real post-training update, added to
every attention and MLP matrix. Each pair uses the same prompt and token
limit: the base model an 8-shot plain-text prompt, the instruct model its chat
template. Differences in points with 95\% paired bootstrap intervals.}
\label{tab:damage}
\begin{center}
\small
\begin{tabular}{lccc}
\toprule
model & accuracy & difference & $|$change$|$ in eRank($W$) \\
\midrule
base                                           & 59.5 & --                         & -- \\
base + Gaussian update, draw 1                 & 4.1  & $-55.4$ ($-58.2$, $-52.7$) & 0.50\% \\
base + Gaussian update, draw 2                 & 4.8  & $-54.7$ ($-57.5$, $-52.2$) & --     \\
base + real singular values, random directions & 5.8  & $-53.8$ ($-56.6$, $-50.9$) & 0.48\% \\
\midrule
instruct (= base + real update)                & 64.5 & --                         & 0.13\% \\
instruct + Gaussian update                     & 11.9 & $-52.6$ ($-55.3$, $-49.9$) & -- \\
\bottomrule
\end{tabular}
\end{center}
\end{table}

An unchanged effective rank also says nothing about whether the model still
works. Adding to Qwen3-0.6B base a random update of the same size as its real
post-training update, matrix by matrix, moves its effective rank by 0.5\%, but
its GSM8K accuracy falls from 59.5\% to 4--6\%, whether the update is Gaussian
or has exactly the real update's singular values (Table~\ref{tab:damage};
details in Appendix~\ref{app:damage}).
The same random update added to the instruct model takes it from 64.5\% to
11.9\%. The real update, which moves the effective rank even less (0.13\%),
produces the instruct model. So an update can wreck a model or turn it into a
working assistant and leave the effective rank of $W$ almost the same either
way. Perplexity on wikitext rises from 18.1 to 33.1 with either random update
(Figure~\ref{fig:random}, right), but the real update raises it too, to 50.1,
because an instruct model no longer predicts plain web text well. An unchanged effective rank therefore cannot tell a
superficial change from a deep one, or from a destructive one. This does not
show that post-training is deep; it shows that this measurement cannot decide
the question.

\subsection{The edit does not need the effective-rank cutoff}
\label{sec:cutoff}

\begin{figure}[t]
\begin{center}
\includegraphics[width=\linewidth]{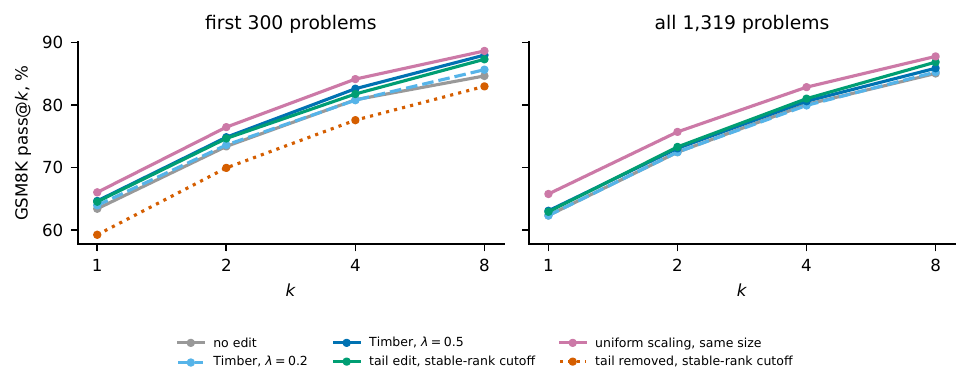}
\end{center}
\caption{GSM8K pass@$k$ for Qwen3-0.6B (non-thinking mode) with tail edits
and with uniform scaling of the whole update. Eight samples
per problem, the same sampling seed per problem for every variant. Left: first
300 test problems, main variants (every variant is in Table~\ref{tab:cutoff300}). Right: all
1{,}319 (Table~\ref{tab:cutoff}).}
\label{fig:cutoff}
\end{figure}

\begin{table}[t]
\caption{Edits of the Qwen3-0.6B update on all 1{,}319 GSM8K test problems;
pass@$k$ in percent.
``Unscaled'' is the share of the update's energy the edit leaves at full
strength (uniform scaling shrinks every direction by the same factor);
``answered'' is the share of samples with a final boxed answer; ``same
size'' means scaled to the size of the stable-rank edit. Differences
from the unedited model in points, with 95\% paired bootstrap intervals over
problems, not corrected for multiple comparisons.}
\label{tab:cutoff}
\begin{center}
\small
\setlength{\tabcolsep}{4pt}
\begin{tabular}{lcccccc}
\toprule
edit & unscaled & pass@1 & pass@8 & answered & $\Delta$ pass@1 & $\Delta$ pass@8 \\
\midrule
none                                       & 100\%  & 62.3 & 85.1 & 97.4\% & -- & -- \\
tail $\times 0.2$, eRank cutoff (Timber)   & 98.9\% & 62.3 & 85.3 & 97.1\% & $-0.0$ ($-0.8$, $+0.8$)  & $+0.2$ ($-1.4$, $+1.8$) \\
tail $\times 0.5$, eRank cutoff (Timber)   & 98.9\% & 63.1 & 85.9 & 97.1\% & $+0.7$ ($-0.0$, $+1.6$)  & $+0.8$ ($-0.6$, $+2.3$) \\
tail $\times 0.5$, stable-rank cutoff      & 41.4\% & 62.9 & 86.9 & 96.0\% & $+0.6$ ($-0.5$, $+1.7$) & $+1.8$ ($+0.2$, $+3.5$) \\
whole update $\times 0.75$ (same size)     & 0\%    & 65.8 & 87.8 & 96.4\% & $+3.4$ ($+2.4$, $+4.5$) & $+2.7$ ($+1.1$, $+4.2$) \\
\bottomrule
\end{tabular}
\end{center}
\end{table}

Because the effective rank of $\dW$ sits close to its noise value, the cutoff
it gives is high: on Qwen3-0.6B it keeps 83\% of the directions and 98.9\% of
the update's energy, so the edit touches only the last 1\% (Wu et al.\ report
98.82\% for Qwen3-8B). An edit that small is also fragile in practice: running
the edited model in bf16 distorts it by 16\% of its size at $\lambda=0.2$ and
25\% at $\lambda=0.5$ (Appendix~\ref{app:cutoff}).

We compare edits of the same update on two benchmarks: GSM8K
\citep{cobbe2021gsm8k}, which is not among Wu et al.'s, and GPQA-Diamond
\citep{rein2024gpqa}, which is. On GSM8K we score with pass@$k$
\citep{chen2021codex} and compare:
Timber's tail edit, the same edit with a stable-rank cutoff that scales 53
times more of the update's energy, and simply scaling the whole update by one
factor per matrix (0.75 on average) so that the edited update has the same
size as the stable-rank edit. We run Timber's edit both at its recommended
$\lambda=0.2$ and at $\lambda=0.5$, and use Qwen3's non-thinking mode, the
mode of Wu et al.'s main table.

On all 1{,}319 problems, Timber's edit has no detectable effect at either
setting (Table~\ref{tab:cutoff}, Figure~\ref{fig:cutoff}). At $\lambda=0.2$
pass@1 is unchanged ($-0.0$ points; $-0.8$, $+0.8$) and pass@8 moves by
$+0.2$ ($-1.4$, $+1.8$); at $\lambda=0.5$ the changes are $+0.7$ ($-0.0$,
$+1.6$) and $+0.8$ ($-0.6$, $+2.3$). The stable-rank edit gives a similar
pass@1 and a pass@8 gain of $+1.8$ ($+0.2$, $+3.5$).

Simply scaling the whole update does better than either tail edit: pass@1
rises by $+3.4$ points ($+2.4$, $+4.5$) and pass@8 by $+2.7$ ($+1.1$,
$+4.2$), with nearly the same share of answered samples (96.4\% against
97.4\%). It makes the update as small as the stable-rank edit does, but evenly
from every direction instead of from the tail. So on this benchmark the gain
comes from making the update smaller, and choosing which directions to shrink
by a spectral cutoff, whether eRank or stable rank, adds nothing. Task
arithmetic already treats the scale of an update as a knob worth tuning
\citep{ilharco2023task}; here that knob alone beats the spectral edits.

\begin{table}[t]
\caption{The same edits on GPQA-Diamond \citep{rein2024gpqa}, Wu et al.'s
benchmark, 198 questions, non-thinking mode with their sampling settings.
``Accuracy'' is the mean over samples (their Mean@$k$), in percent, as is
pass@4; eight samples per
question for the unedited model, four for each edit. Differences in points
with 95\% paired bootstrap intervals over questions.}
\label{tab:gpqa}
\begin{center}
\small
\setlength{\tabcolsep}{4pt}
\begin{tabular}{lcccc}
\toprule
edit & accuracy & $\Delta$ accuracy & pass@4 & $\Delta$ pass@4 \\
\midrule
none                                      & 25.9 & --                       & 46.6 & -- \\
tail $\times 0.2$, eRank cutoff (Timber)  & 26.9 & $+0.9$ ($-2.1$, $+4.0$)  & 46.0 & $-0.6$ ($-5.9$, $+4.8$) \\
tail $\times 0.5$, eRank cutoff (Timber)  & 26.4 & $+0.4$ ($-2.5$, $+3.5$)  & 47.5 & $+0.9$ ($-4.2$, $+6.4$) \\
tail $\times 0.8$, eRank cutoff (Timber)  & 26.9 & $+0.9$ ($-2.0$, $+3.9$)  & 47.0 & $+0.4$ ($-4.8$, $+5.6$) \\
tail $\times 0.2$, stable-rank cutoff     & 21.3 & $-4.6$ ($-9.3$, $+0.2$)  & 48.0 & $+1.4$ ($-6.4$, $+9.4$) \\
whole update, scaled to the stable-rank edit's size & 22.9 & $-3.1$ ($-7.1$, $+0.9$)  & 54.0 & $+7.5$ ($+0.5$, $+14.3$) \\
\bottomrule
\end{tabular}
\end{center}
\end{table}

GPQA-Diamond is where Wu et al.\ report their largest gain for this model: a
mean accuracy of 28.79\% against 23.74\% unedited, $+5.1$ points. Our
unedited model scores 25.9\%, about two points above theirs. Timber's edit, at
each of the three $\lambda$ values they searched, moves accuracy by under a
point, and every interval includes zero (Table~\ref{tab:gpqa}). We do not
observe their gain, but our setup is our own, not their evaluation harness:
our prompt and answer format, our answer extraction, four samples per
question against their three full repeats, and 2{,}048 new tokens against
their 8{,}192. Truncation, at least, does not explain the difference. The
share of samples that end without a final answer is 3.5\% for the unedited
model and 3.4--4.2\% for Timber's three edits (8.7--8.8\% for the two larger
edits), so Timber's edits are cut off no more often than no edit. A rerun of
the unedited model and the recommended $\lambda=0.2$ with the limit doubled
to 4{,}096 tokens, the most this model's samples fit on our GPU, gives 25.6\%
against 25.4\%, a difference of $-0.3$ points ($-3.5$, $+3.0$), with
2.7--2.9\% of samples unfinished (Appendix~\ref{app:cutoff}). With 198
questions our intervals are about $\pm3$ points, and their $+5.1$, from the
same 198 questions, carries a similar uncertainty. Under our settings the
intervals exclude a gain of five points; a gain of one to three points is not
ruled out. Here, unlike on GSM8K, the larger edits hurt: shrinking most of
the update, whether by the stable-rank cutoff or
evenly, costs 3--5 points of mean accuracy (intervals just reaching zero) and
makes the model answer at length, while the uniform version's pass@4 rises.
That pattern, worse single answers but more varied ones, is a change in
behaviour, not a clear improvement, and with about thirty comparisons across
our edit tables one interval just clearing zero is what chance would give.

Taken together, the two benchmarks say the same thing about the cutoff. An
edit that scales only the last 1\% of the update's energy does nothing we can
measure, on GSM8K or on GPQA. Edits that scale a quarter of the update change
the model's behaviour much more, for better (GSM8K) or for worse (GPQA), and
the spectral cutoff is not what decides it.

On the first 300 problems we also ran the extreme $\lambda=0$, which removes
the tail completely (Table~\ref{tab:cutoff300}). Removing Timber's tail, 1.1\%
of the update's energy, changes nothing measurable ($+0.5$ points of pass@1).
Removing the stable-rank ``tail'', which is 59\% of the update's energy and so
most of the update, costs 4.2 points against the unedited model and 5.3 against
halving it. Removing most of an update hurts; this says little about the
weakest directions alone. A uniform scaling matched in size to Timber's edit is
almost erased by bf16 storage (88\% of the change is lost) and, like Timber's
edit, has no clear effect ($+0.8$; $-0.8$, $+2.4$).

We did not test thinking mode, where Wu et al.\ also report gains, and our
GPQA intervals are wide. Within those limits, the effective-rank cutoff is not
what makes an edit of this kind work.

%% file: sections/related.tex
\section{Related work}
\label{sec:related}

\paragraph{Noise baselines for spectra.} Comparing a spectrum against what
noise would give is old. Parallel analysis \citep{horn1965parallel} keeps only
the factors whose eigenvalues beat those of random data, and
\citet{gavish2014optimal} derive the optimal threshold for separating signal
singular values from Gaussian noise. Random matrix theory has also been used to
study trained weight matrices: their spectra depart from those of random
matrices in heavy-tailed ways that track training quality
\citep{martin2021implicit}, while much of the bulk still follows random-matrix
statistics \citep{thamm2022random}. \citet{staats2023boundary} use those
statistics to separate the noise part of a weight matrix from its information,
and \citet{hartford2024spectrum} use them to choose which layers to train. For
fine-tuning, \citet{biderman2024lora}
compared the spectrum of a fully fine-tuned weight matrix with that of the
pretrained matrix plus Gaussian noise and found them alike. We compare the
update itself with noise, measure by measure, and find that the answer depends
on which measure one uses.

\paragraph{Superficial alignment.} \citet{zhou2023lima} proposed that
alignment mostly selects a format the pretrained model already supports, and
\citet{lin2024urial} found that base and aligned models differ mainly on
stylistic tokens. Others have questioned how far this goes: safety behaviour
can be confined to the first few output tokens \citep{qi2025safety}, and
post-training keeps improving reasoning with more data
\citep{raghavendra2024revisiting}. \citet{wu2025timber} looked for
weight-level evidence through the effective rank of released models
(\S\ref{sec:timber}). \citet{chaudhury2025alignment} locates the effect of
preference tuning in a few middle layers by activation patching, a question
about activations that our weight analysis does not address.

\paragraph{The rank of fine-tuning updates.} Fine-tuning can be carried out in
a small random subspace \citep{aghajanyan2021intrinsic}, the idea behind LoRA
\citep{hu2022lora}, and fine-tuned models can be stored as low-rank
approximations of the update; \citet{ryu2023efficient} also plot the singular
values of an update against a Marchenko--Pastur reference and find them more
concentrated. \citet{zhang2023tiny} find that fine-tuning moves the
weights within a small task-specific subspace. Measured directly, full
fine-tuning updates need a rank 10--100 times larger than typical LoRA ranks to
hold 90\% of their energy \citep{biderman2024lora}, and reinforcement-learning
and preference-tuning updates are reported as nearly full rank, 99.2--99.8\%
of the maximum
\citep{mukherjee2025rl}. Both statements count directions; our results agree
with them and add that the same updates are far more concentrated than noise at
the top of the spectrum. \citet{shuttleworth2025lora} compare singular vectors
of fine-tuned and pretrained $W$ and find that LoRA introduces new dominant
directions that full fine-tuning does not. Delta-CoMe \citep{ping2024deltacome}
observes that update spectra are long-tailed and compresses them by giving more
bits to the top singular vectors, finding that plain low-rank truncation hurts;
BitDelta \citep{liu2024bitdelta} goes further and keeps only the sign of each
entry of the update plus one scale per matrix, with little loss.
\citet{paul2026spectral} uses stable rank among other features of LoRA updates
to tell training objectives apart. Closest to our head and tail,
\citet{zhang2026preference} describe a sharp head and a residual tail in the
updates of DPO and GRPO runs with LoRA and study what each part does to the
model's behaviour; we ask how such spectra compare with noise and with
fine-tuning that has no preferences to learn.

\paragraph{Editing updates.} Updates can be added, scaled and merged
\citep{ilharco2023task}, and much of an update can be dropped or trimmed at
random or by magnitude before merging with little loss
\citep{yu2024dare, yadav2023ties}. Scaling an update as a whole is an old and
effective edit: interpolating between pretrained and fine-tuned weights makes
fine-tuning more robust \citep{wortsman2022wiseft}, and extrapolating along
the alignment update can improve an aligned model \citep{zheng2025expo}.
Timber's tail edit belongs to this family, and our comparison with uniform
scaling (\S\ref{sec:cutoff}) asks whether its spectral cutoff adds anything
over the simplest edit of this kind.

\paragraph{Low precision.} Rounding weights to bf16 cancels small updates
\citep{zamirai2020bfloat16}, and in reinforcement learning this can make
updates look sparse \citep{zhu2025path}. We found the same effect in our own
first runs (\S\ref{sec:setup}) and show that it can also inflate the
threshold count of released checkpoints with small updates
(\S\ref{sec:released}).

\paragraph{Why updates concentrate.} Gradient descent spends most of its
motion in a small subspace \citep{gurari2018tiny}, and gradients of linear
layers are approximately low rank \citep{zhao2024galore}, so a sum of gradient
steps can inherit a few dominant directions. We do not test this mechanism.

%% file: sections/discussion.tex
\section{Discussion}
\label{sec:discussion}

\paragraph{What the measures are good for.} None of the five measures is
wrong; they answer different questions. Counting measures say how many
directions an update touches, and for fine-tuning the answer is ``almost
all''. Stable rank and the participation ratio say how unevenly the energy is
spread, and the answer is ``very''. We do not recommend stable rank over the
others: it depends on a single direction, it is also far below noise for the
pretrained weights and for the difference of two unrelated checkpoints, not
only for a real update, and a cutoff built on it edited a model no better than
one built on eRank. The mistake is to read any one of these numbers without the value for
noise next to it, or to take a property of all fine-tuning as evidence about
alignment.

\paragraph{Check where a checkpoint comes from.} The obvious base for the
OLMo 2 1B SFT checkpoint, the main revision of the released base model, is not
its parent: the two share no initialisation, and the real parent is the final
mid-training checkpoint, published only as a branch (\S\ref{sec:setup}). The
difference between the unrelated pair still has a plausible-looking spectrum,
with stable rank 4.4 times below noise, much like the real update. Nothing in
a rank measure flags such a mistake; a cosine between the two sets of weights
does, and costs nothing.

\paragraph{Limitations.}
\begin{itemize}
  \item \textbf{One model for preference training.} The DPO runs and their
  swapped control are on Qwen2.5-0.5B only. Reinforcement learning appears only
  in the released OLMo 2 stage, which is too small to measure above rounding.
  \item \textbf{Scale of our own training.} Our fine-tunes are 0.36B and 0.5B
  parameters with about two million tokens per epoch. The measurement results
  hold on released models up to 8B, but we do not study training at that scale.
  \item \textbf{Tokens or steps.} Matching the training tokens leaves the
  code runs with 2.6--2.7 times fewer optimizer steps than SFT; matching the steps
  gives them 2.5--2.7 times more tokens. No single control matches both, and the
  two give different code updates.
  \item \textbf{Which null.} A Gaussian matrix is the least structured baseline
  possible, and even the difference of two unrelated trained checkpoints beats
  it. We add the pretrained weights as a harder reference, and matched
  fine-tunes without alignment content as the real test, but other baselines
  (for instance updates from other optimizers) could tell more.
  \item \textbf{Seeds.} The SFT, wikitext, code and DPO runs were repeated
  once; the shuffled-words and step-matched runs were not. Two seeds bound
  the run-to-run variation only roughly.
  \item \textbf{Released models are black boxes.} Apart from OLMo 2, their
  data and recipes differ in many ways at once, so they test the measurement,
  not any claim about alignment.
  \item \textbf{Loss masking differs.} SFT computes the loss on answer tokens
  only, the controls on every token; the tokens seen are matched, the tokens
  trained on are not.
  \item \textbf{The shuffled-words result is unexplained.} The one explanation
  we tested, alignment with token frequency, is shared by other runs.
  \item \textbf{Directions, measured coarsely.} The overlap of top-16
  subspaces shows that the data steers the directions of an update; it does
  not say which directions, or what they do, and it ignores their sign. The
  signed cosine separates real from swapped DPO, but with shared training pairs
  and one swapped run it cannot say how much of that is preference learning.
  \item \textbf{The edit experiment is narrow:} one model, non-thinking mode
  only, and 198 GPQA questions, which leaves intervals of about $\pm3$
  points.
  \item \textbf{One random-update test.} The damage test uses one model,
  Qwen3-0.6B, and one task, GSM8K.
\end{itemize}

\section{Conclusion}
\label{sec:conclusion}

Rank measures that count directions describe fine-tuning updates as nearly
full rank and put them only a little below random matrices of the same shape.
Measures
that weigh the top directions show the same updates to be far more
concentrated than noise, and more concentrated than the pretrained weights.
Both are right: the updates have a few strong directions and a long tail that
holds most of the energy. With the amount of training held fixed, the data
shapes an update somewhat, but instruction tuning does not stand out from
Wikipedia or code; DPO gives the same shape with real or half-swapped
preferences; and scrambled text gives the most concentrated
update. The shape is a property of fine-tuning, not a mark of alignment. An
unchanged effective rank of the weights after post-training is
what a random update of the same size also produces, and an edit built on
the effective rank of the update does nothing measurable on two benchmarks.
Our
recommendation is simple: report any rank measure of a weight update next to
its value on a random matrix of the same shape, check that low-precision
training or storage did not erase the update, check that the two checkpoints
are really related, and test claims about alignment against a fine-tune
without alignment content.

\subsubsection*{Broader Impact Statement}
This paper is about measurement methodology for analysing fine-tuned language
models; it introduces no new capabilities or training methods. Its main
practical consequence is caution about reading weight-space statistics as
evidence that alignment is shallow or deep, since such claims can affect how
much trust is placed in alignment training.

\subsubsection*{Reproducibility}
Every run records its configuration, its exact training rows, its seed, its
token count and the fraction of weights left unchanged. The code, the
configurations and run manifests, the saved singular-value spectra of every
update, the random-update seeds, the edit scripts and the evaluation prompts
are at \url{https://github.com/usman7384/llm-finetuning-update-spectra}.

%% file: sections/appendix.tex
\section{Training details}
\label{app:training}

\begin{table}[H]
\caption{Settings shared by all our fine-tunes. Each language-model run trains
on the same number of tokens as the SFT run of the same model, counted after
truncation. DPO uses the same precision, memory settings and batch, with the
changes listed at the bottom.}
\label{tab:hparams}
\begin{center}
\small
\begin{tabular}{ll}
\toprule
setting & value \\
\midrule
weights / compute precision & fp32 master weights, bf16 autocast \\
optimizer                   & 8-bit AdamW, learning rate $2\times10^{-5}$, no weight decay \\
schedule                    & cosine, 3\% warmup, 3 epochs, gradient clipping at 1.0 \\
batch                       & 1 sequence $\times$ 16 accumulation steps \\
sequence length             & 512 tokens \\
embeddings                  & frozen (tied with the output head; not analysed) \\
memory                      & gradient checkpointing, one 6\,GB GPU \\
check after training        & at most $1.01\times10^{-5}$ of attention/MLP weights unchanged \\
\midrule
SFT data                    & 10k examples of Alpaca-cleaned, loss on answers only \\
wikitext / code data        & wikitext-103 / CodeParrot (Python), plain LM loss \\
shuffled-words data         & wikitext documents, word order shuffled within each document \\
training tokens per epoch   & Qwen: 1.71--1.75M; SmolLM2: 1.87--1.91M \\
\midrule
DPO (Qwen only)             & from the Qwen SFT run; 10k UltraFeedback pairs, 1 epoch, \\
                            & learning rate $5\times10^{-6}$, $\beta=0.1$, reference log-probs precomputed \\
swapped-DPO control         & same pairs, chosen and rejected swapped in a random half \\
\bottomrule
\end{tabular}
\end{center}
\end{table}

\begin{table}[H]
\caption{Every fine-tune. Steps are optimizer steps (rounded to 10); loss is
the mean training loss; $\|\dW\|_F$ is over all attention and MLP matrices.
The second SFT seed was interrupted and resumed from step 1{,}600 with its
optimizer and scheduler state, so its logged mean loss covers only the last
steps and is not shown.}
\label{tab:runs}
\begin{center}
\small
\begin{tabular}{llcccccc}
\toprule
model & run & steps & loss & $\|\dW\|_F$ & stable rank & stable w/o top & hours \\
\midrule
Qwen2.5-0.5B & SFT               & 1{,}840 & 1.00 & 10.4 & 0.097 & 0.126 & 2.8 \\
             & SFT, seed 2       & 1{,}840 & --   & 10.5 & 0.096 & 0.126 & -- \\
             & wikitext          & 1{,}910 & 2.23 & 12.1 & 0.093 & 0.122 & 2.8 \\
             & wikitext, seed 2  & 1{,}910 & 2.24 & 12.0 & 0.092 & 0.121 & 2.7 \\
             & code              & 680     & 1.11 & 6.3  & 0.094 & 0.123 & 1.3 \\
             & code, seed 2      & 670     & 1.12 & 6.3  & 0.094 & 0.122 & 1.3 \\
             & shuffled words    & 1{,}920 & 5.40 & 12.3 & 0.042 & 0.072 & 2.9 \\
             & code, steps matched & 1{,}820 & 1.13 & 10.1 & 0.114 & 0.141 & 3.5 \\
             & DPO               & 570     & 0.69 & 1.37 & 0.053 & 0.088 & 6.7 \\
             & DPO, seed 2       & 570     & 0.70 & 1.37 & 0.054 & 0.089 & 3.6 \\
             & swapped DPO       & 570     & 0.83 & 1.37 & 0.055 & 0.093 & 8.6 \\
\midrule
SmolLM2-360M & SFT               & 1{,}840 & 1.18 & 14.9 & 0.025 & 0.041 & 3.4 \\
             & SFT, seed 2       & 1{,}840 & 1.17 & 14.8 & 0.025 & 0.041 & 3.2 \\
             & wikitext          & 2{,}080 & 2.52 & 17.4 & 0.020 & 0.032 & 3.7 \\
             & wikitext, seed 2  & 2{,}060 & 2.52 & 17.2 & 0.020 & 0.032 & 3.6 \\
             & code              & 720     & 1.28 & 8.8  & 0.034 & 0.051 & 1.4 \\
             & code, seed 2      & 720     & 1.28 & 8.8  & 0.034 & 0.051 & 1.3 \\
             & shuffled words    & 2{,}090 & 5.51 & 20.8 & 0.009 & 0.021 & 3.7 \\
             & code, steps matched & 1{,}800 & 1.26 & 14.6 & 0.042 & 0.060 & 3.2 \\
\bottomrule
\end{tabular}
\end{center}
\end{table}

Training a run took 1.3--3.7 hours on one GPU, and 3.6--8.6 hours for DPO
(Table~\ref{tab:runs}). Held-out reward accuracy, on about 190 validation
pairs that are never swapped, is 65\% and 74\% for the two DPO seeds and 51\%
for the swapped control. The base
tokenizer of SmolLM2-360M has no chat template, so SFT borrows the one from
SmolLM2-360M-Instruct. Second seeds change both the training seed and the data
sample. Every run records its exact training rows, token count and the
fraction of unchanged weights. Dropping the top direction leaves every update
far below noise, whose stable rank without its own top direction is 0.440 for
Qwen and 0.357 for SmolLM2, so stable rank is not driven by a single outlying
direction.

\section{Threshold sensitivity and the noise baseline}
\label{app:sensitivity}

\begin{table}[H]
\caption{The threshold count at three cutoffs. At $0.01\,\sigma_1$ every
update looks almost like noise; at $0.1\,\sigma_1$ the gap is large. The verdict
depends on an arbitrary choice.}
\label{tab:threshold}
\begin{center}
\small
\begin{tabular}{llccc}
\toprule
model & matrix & $0.01\,\sigma_1$ & $0.05\,\sigma_1$ & $0.1\,\sigma_1$ \\
\midrule
Qwen2.5-0.5B & SFT update            & 0.973 & 0.889 & 0.787 \\
             & wikitext update       & 0.972 & 0.886 & 0.775 \\
             & code update           & 0.975 & 0.895 & 0.799 \\
             & shuffled-words update & 0.941 & 0.752 & 0.530 \\
             & noise                 & 0.996 & 0.982 & 0.964 \\
             & pretrained $W$        & 0.976 & 0.913 & 0.851 \\
\midrule
SmolLM2-360M & SFT update            & 0.951 & 0.757 & 0.458 \\
             & wikitext update       & 0.948 & 0.701 & 0.386 \\
             & code update           & 0.963 & 0.803 & 0.534 \\
             & shuffled-words update & 0.863 & 0.384 & 0.151 \\
             & noise                 & 0.996 & 0.982 & 0.964 \\
             & pretrained $W$        & 0.982 & 0.926 & 0.857 \\
\bottomrule
\end{tabular}
\end{center}
\end{table}

\begin{table}[H]
\caption{Stable rank of the random-init pair against the prediction
$mn/(\sqrt{m}+\sqrt{n})^2/r$ for a Gaussian matrix of the same shape,
Qwen2.5-0.5B, mean over layers.}
\label{tab:mp}
\begin{center}
\small
\begin{tabular}{lccc}
\toprule
matrix & shape & predicted & random-init pair \\
\midrule
\texttt{q}, \texttt{o} & $896 \times 896$  & 0.250 & 0.252, 0.251 \\
\texttt{k}, \texttt{v} & $128 \times 896$  & 0.527 & 0.536, 0.535 \\
\texttt{gate}, \texttt{up} & $4864 \times 896$ & 0.490 & 0.492, 0.492 \\
\texttt{down} & $896 \times 4864$ & 0.490 & 0.493 \\
\bottomrule
\end{tabular}
\end{center}
\end{table}

\begin{figure}[H]
\begin{center}
\includegraphics[width=\linewidth]{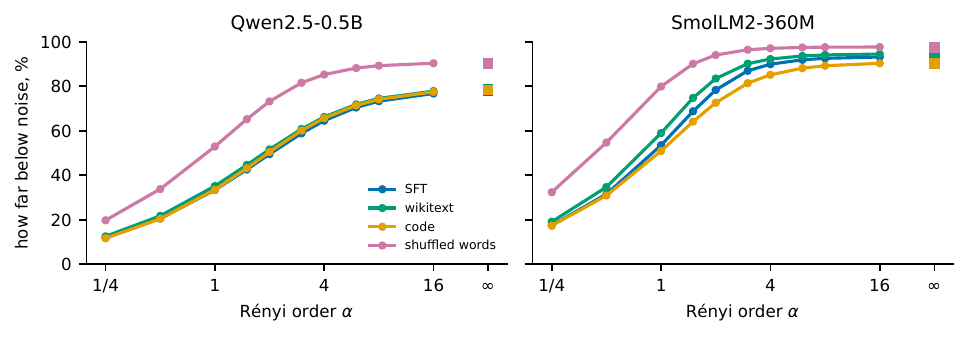}
\end{center}
\caption{How far below noise each update is, $1 - R_\alpha(\text{update}) /
R_\alpha(\text{noise})$ as means over matrices, along the R\'enyi family from
$\alpha = 1/4$ to $\alpha = \infty$ (stable rank, squares). The gap grows with
the order and levels off from about $\alpha = 8$.}
\label{fig:renyi}
\end{figure}

\section{Stable rank by matrix type}
\label{app:pertype}

\begin{table}[H]
\caption{Stable rank (fraction of maximum rank) by matrix type.}
\label{tab:pertype}
\begin{center}
\small
\begin{tabular}{llccccccc}
\toprule
model & matrix & \texttt{q} & \texttt{k} & \texttt{v} & \texttt{o} & \texttt{gate} & \texttt{up} & \texttt{down} \\
\midrule
Qwen2.5-0.5B & SFT update            & 0.032 & 0.064 & 0.251 & 0.057 & 0.064 & 0.072 & 0.139 \\
             & wikitext update       & 0.031 & 0.072 & 0.227 & 0.057 & 0.065 & 0.085 & 0.112 \\
             & code update           & 0.031 & 0.070 & 0.243 & 0.058 & 0.057 & 0.073 & 0.127 \\
             & shuffled-words update & 0.011 & 0.043 & 0.124 & 0.021 & 0.023 & 0.026 & 0.048 \\
             & noise                 & 0.252 & 0.536 & 0.535 & 0.251 & 0.492 & 0.492 & 0.493 \\
             & pretrained $W$        & 0.058 & 0.212 & 0.453 & 0.078 & 0.071 & 0.242 & 0.135 \\
\midrule
SmolLM2-360M & SFT update            & 0.017 & 0.045 & 0.032 & 0.010 & 0.021 & 0.020 & 0.028 \\
             & wikitext update       & 0.015 & 0.037 & 0.029 & 0.010 & 0.018 & 0.016 & 0.018 \\
             & code update           & 0.023 & 0.047 & 0.051 & 0.022 & 0.023 & 0.026 & 0.049 \\
             & shuffled-words update & 0.005 & 0.014 & 0.013 & 0.005 & 0.008 & 0.007 & 0.013 \\
             & noise                 & 0.251 & 0.407 & 0.407 & 0.251 & 0.387 & 0.386 & 0.387 \\
             & pretrained $W$        & 0.056 & 0.128 & 0.312 & 0.086 & 0.087 & 0.207 & 0.106 \\
\bottomrule
\end{tabular}
\end{center}
\end{table}

Every matrix type of every update is far below noise in both models. Within a
model the SFT, wikitext and code updates are close in every column, and the
shuffled-words update is the most concentrated in every column. On SmolLM2
every type of every update is also below the pretrained weights.

\section{The shuffled-words update}
\label{app:shuffled}

\begin{table}[H]
\caption{Stable rank by quarter of the network.}
\label{tab:shuffled}
\begin{center}
\small
\begin{tabular}{llcccc}
\toprule
model & update & first quarter & second & third & last quarter \\
\midrule
Qwen2.5-0.5B & SFT            & 0.094 & 0.099 & 0.103 & 0.091 \\
             & wikitext       & 0.086 & 0.100 & 0.104 & 0.080 \\
             & code           & 0.091 & 0.097 & 0.104 & 0.084 \\
             & shuffled words & 0.055 & 0.039 & 0.034 & 0.041 \\
\midrule
SmolLM2-360M & SFT            & 0.023 & 0.027 & 0.029 & 0.021 \\
             & wikitext       & 0.019 & 0.024 & 0.023 & 0.015 \\
             & code           & 0.030 & 0.038 & 0.043 & 0.027 \\
             & shuffled words & 0.013 & 0.010 & 0.007 & 0.007 \\
\bottomrule
\end{tabular}
\end{center}
\end{table}

We tested one possible explanation: that the shuffled-words update mainly
shifts the output towards frequent tokens. For each layer's
\texttt{down} projection update on Qwen, we map the top left singular vector to the
vocabulary through the final norm and the tied unembedding (ignoring the
layers in between, as the logit lens does) and correlate the resulting logit
shift with the log frequency of each token in the run's training data,
against 50 random directions. In the last third of the network the top
direction of several runs lines up with frequency, not only shuffled words:
the correlation beats the random 95th percentile (about 0.19) in layers 18--23
for shuffled words, in most of layers 15--23 for SFT and code, and in layer 23
for wikitext. The mean correlation over the last third is 0.33 for shuffled
words, 0.25 for SFT, 0.19 for code and 0.09 for wikitext. With 24 layers and
four runs, and no correction for multiple tests, we read this only as a
pattern: frequency alignment in late layers is common to fine-tuning, so it
does not explain why the shuffled-words update is the most concentrated.

\section{Released pairs: more measures}
\label{app:released}

\begin{table}[H]
\caption{More on the released pairs. ``Particip.\ ratio'': the participation
ratio, next to its value for a Gaussian matrix of the same shape. ``Above
edge'': the number of singular values above the largest one such a Gaussian
matrix would have (mean per matrix). ``Rounding share'': the number of
singular values of the simulated bf16 rounding error above $0.01\,\sigma_1$ of
the update, as a share of the update's threshold count; a loose upper bound on
how much of the count rounding could explain, as a ratio of means over
matrices (compare the ``above floor'' column of Table~\ref{tab:released}).
$\|\dW\|_F/\|W\|_F$: the size of the update relative to the base weights.
``Zero entries'': the share of entries of $\dW$ that are exactly zero, the
direct trace of rounding.}
\label{tab:released_more}
\begin{center}
\small
\setlength{\tabcolsep}{4pt}
\begin{tabular}{lcccccc}
\toprule
update & particip.\ ratio & noise & above edge & rounding share & $\|\dW\|_F/\|W\|_F$ & zero entries \\
\midrule
SmolLM2-135M instruct & 0.321 & 0.667 & 30  & 28\% & 8.1\%  & 2.5\%  \\
SmolLM2-360M instruct & 0.239 & 0.668 & 46  & 10\% & 7.6\%  & 2.7\%  \\
Qwen2.5-0.5B instruct & 0.657 & 0.755 & 22  & 76\% & 7.8\%  & 2.2\%  \\
Qwen3-0.6B            & 0.442 & 0.654 & 72  & 15\% & 12.7\% & 1.5\%  \\
Qwen2.5-1.5B instruct & 0.299 & 0.753 & 100 & 94\% & 1.2\%  & 14.0\% \\
SmolLM2-1.7B instruct & 0.375 & 0.629 & 140 & 45\% & 6.2\%  & 3.1\%  \\
Qwen3-8B              & 0.405 & 0.693 & 243 & 24\% & 8.4\%  & 2.1\%  \\
OLMo 2 1B: base $\to$ SFT & 0.381 & 0.629 & 137 & --   & 9.8\% & 0.0\%  \\
OLMo 2 1B: SFT $\to$ DPO  & 0.116 & 0.629 & 114 & 98\% & 0.4\% & 38.8\% \\
OLMo 2 1B: DPO $\to$ RL   & 0.127 & 0.629 & 99  & 100\% & 0.3\% & 51.6\% \\
Qwen2.5-Coder-0.5B    & 0.545 & 0.755 & 44  & 0\% & 89\%   & 0.2\%  \\
Qwen2.5-Math-1.5B     & 0.611 & 0.753 & 61  & 0\% & 156\%  & 0.1\%  \\
Qwen2.5-Coder-1.5B    & 0.585 & 0.753 & 66  & 0\% & 92\%   & 0.1\%  \\
\bottomrule
\end{tabular}
\end{center}
\end{table}

Qwen2.5-0.5B is the least concentrated instruct release on every rank measure
(not on top-1 energy), and less concentrated than our own SFT of the same
base; we do not know why.

The two direct checks agree. Where few directions rise above the rounding
floor, many entries of the update are exactly zero: 14\% for Qwen2.5-1.5B, and
39\% and 52\% for the OLMo DPO and RL stages, whose updates are under half a
percent of the weights. For the other instruct releases 1.5--3.1\% of entries
are zero and 91--99\% of directions are above the floor. The rounding share is much higher for several
of these pairs (for instance 76\% for Qwen2.5-0.5B), which shows how loose a
bound it is: it compares rounding with $0.01\,\sigma_1$, not with the
directions the update actually has. For the specialists, whose updates are as
large as the weights themselves, almost no entry is zero.

\section{Direction overlap between updates}
\label{app:overlap}

For each matrix we take the top-16 left singular vectors of each of two
updates, $U_A$ and $U_B$, and compute $\|U_A^\top U_B\|_F^2 / 16$, which is 1
when the two subspaces coincide and $16/m$ for two random 16-dimensional
subspaces of $\R^m$. We report the mean over matrices with a layer-bootstrap
interval (Table~\ref{tab:overlap}). Right singular vectors give the same
ordering, with smaller overlaps between runs on different data.

\begin{table}[H]
\caption{Overlap of the top-16 left singular subspaces of two updates, mean
over matrices, with 95\% layer-bootstrap intervals.}
\label{tab:overlap}
\begin{center}
\small
\begin{tabular}{lcc}
\toprule
pair & Qwen2.5-0.5B & SmolLM2-360M \\
\midrule
SFT vs SFT, seed 2                & 0.32 (0.30, 0.33) & 0.51 (0.50, 0.52) \\
wikitext vs wikitext, seed 2      & 0.30 (0.29, 0.31) & 0.49 (0.47, 0.52) \\
code vs code, seed 2              & 0.26 (0.25, 0.27) & 0.34 (0.33, 0.36) \\
code vs code, steps matched       & 0.37 (0.36, 0.39) & 0.54 (0.53, 0.55) \\
\midrule
SFT vs wikitext                   & 0.22 (0.21, 0.23) & 0.13 (0.13, 0.14) \\
SFT vs code                       & 0.20 (0.19, 0.21) & 0.12 (0.12, 0.12) \\
wikitext vs code                  & 0.20 (0.19, 0.21) & 0.12 (0.12, 0.14) \\
SFT vs shuffled words             & 0.17 (0.16, 0.18) & 0.10 (0.10, 0.11) \\
wikitext vs shuffled words        & 0.21 (0.20, 0.23) & 0.18 (0.16, 0.20) \\
\midrule
DPO vs DPO, seed 2                & 0.28 (0.27, 0.29) & -- \\
DPO vs swapped DPO                & 0.27 (0.26, 0.28) & -- \\
\midrule
SFT vs random subspaces           & 0.04              & 0.02 \\
\bottomrule
\end{tabular}
\end{center}
\end{table}

Two things stand out. The data moves the directions more than it moves the
shape: on SmolLM2 the overlap between runs on different data (0.12--0.13) is a
quarter of the overlap between seeds of the same run (0.34--0.51), while their
stable ranks differ by at most 0.014. And wikitext and shuffled words, which
share their words, overlap more than wikitext and code do.

The overlap compares unsigned subspaces on one side of the update, so it is
blind to sign, to how left and right directions pair up, and to how strong each
one is. Table~\ref{tab:signed} gives the cosine between two updates as flat
vectors, $\langle\dW_A,\dW_B\rangle / (\|\dW_A\|_F\,\|\dW_B\|_F)$, averaged
over matrices. Runs on different data, whose subspaces overlap by 0.10--0.22,
have cosines of 0.001--0.006, except wikitext and shuffled words, which are the
same documents. So the overlaps between unrelated data reflect directions that
fine-tuning uses in general, not agreement between the updates. The last column
counts training rows in common. Shared rows can make two updates agree by
themselves, but they need not: the two wikitext seeds share 1\% of their rows
and still have a cosine of 0.08 on Qwen.

\begin{table}[H]
\caption{Signed cosine between two updates, mean over matrices, with 95\%
layer-bootstrap intervals over layers, not over training runs. ``Shared'' is
the share of the first run's training rows that the second run also uses, on
Qwen2.5-0.5B.}
\label{tab:signed}
\begin{center}
\small
\begin{tabular}{lccc}
\toprule
pair & Qwen2.5-0.5B & SmolLM2-360M & shared \\
\midrule
SFT vs SFT, seed 2                & 0.189 (0.181, 0.198) & 0.420 (0.411, 0.430) & 19\% \\
wikitext vs wikitext, seed 2      & 0.080 (0.070, 0.092) & 0.329 (0.302, 0.357) & 1\% \\
code vs code, seed 2              & 0.083 (0.080, 0.086) & 0.337 (0.327, 0.348) & 6\% \\
code vs code, steps matched       & 0.448 (0.433, 0.464) & 0.676 (0.670, 0.681) & 97\% \\
\midrule
SFT vs wikitext                   & 0.006 (0.005, 0.007) & 0.004 (0.003, 0.005) & -- \\
SFT vs code                       & 0.003 (0.002, 0.004) & 0.004 (0.003, 0.005) & -- \\
wikitext vs code                  & 0.004 (0.003, 0.005) & 0.003 (0.002, 0.004) & -- \\
SFT vs shuffled words             & 0.002 (0.001, 0.003) & 0.001 (0.001, 0.002) & -- \\
wikitext vs shuffled words        & 0.068 (0.051, 0.087) & 0.100 (0.084, 0.115) & 100\% \\
\midrule
DPO vs DPO, seed 2                & 0.084 (0.082, 0.087) & -- & 16\% \\
DPO vs swapped DPO                & 0.040 (0.039, 0.042) & -- & 100\% \\
DPO, seed 2 vs swapped DPO        & 0.001 ($-$0.000, 0.002) & -- & 16\% \\
\bottomrule
\end{tabular}
\end{center}
\end{table}

\section{Random-update details}
\label{app:damage}

Each random update is drawn matrix by matrix with its own seed and scaled to
the Frobenius norm of the real update of that matrix; embeddings and norms are
left alone. The edit is made in fp32 and the model then runs in bf16. Answers
are greedy, up to 256 new tokens for the base model (8-shot prompt from the
first eight GSM8K training problems, answer after ``The answer is'') and 512
for the instruct model (chat template, thinking off, answer in
\verb|\boxed{}|). Many answers from the randomly updated base models hit the
token limit (29--59\%, against 5\% for the unedited base), but even the ones
that finish are right only 7--8\% of the time, against 63\% for the base model.
The instruct model with a random update hits its 512-token limit on 72\% of
problems; on the rest it is right 28\% of the time, against 66\% for the
unedited instruct model. Part of its drop to 11.9\% is therefore rambling
rather than wrong answers, but the finished answers show the damage too.

We also scored the base model plus its real update with the base model's
8-shot prompt. It reaches only 19.9\%. That is not clear evidence of damage:
post-trained weights write long chat-style reasoning in that format, and 59\%
of their answers hit the 256-token limit, about as many as with the first
Gaussian update. What separates them is the answers that finish: 42\% are
right with the real update, 7--8\% with the random ones. This combination also
keeps the base model's embeddings, which post-training changed by 13\%. The
fair test of the real update is the instruct model in its own format
(Table~\ref{tab:damage}).

\section{Edit experiment details}
\label{app:cutoff}

On GPQA-Diamond the prompt gives the question and its four options, shuffled
once per question with a fixed seed, and asks for ``Answer: \$LETTER'' as the
last line; the answer is the last such line (or the last \verb|\boxed{}|
letter). We use Wu et al.'s sampling settings for this mode (temperature 0.7,
top-$p$ 0.8, top-$k$ 20), with up to 2{,}048 new tokens where they allow
8{,}192, eight samples per question for the unedited model and four for each
edit where they run three full repeats; pass@4 is the unbiased estimate from
those samples. The share of samples that reach the limit without a final
answer is 3.5\% unedited, 3.8\%, 4.2\% and 3.4\% for Timber's edit at
$\lambda = 0.2$, $0.5$ and $0.8$, and 8.8\% and 8.7\% for the stable-rank
cutoff and uniform scaling, which make the model answer at length.

To check that the limit is not hiding a gain, we reran the unedited model and
the $\lambda = 0.2$ edit with 4{,}096 new tokens, four samples per question
for both. Wu et al.'s 8{,}192 does not fit on our 6\,GB GPU with four samples
at once, because on a few questions this model writes very long answers. We
began the unedited model at that limit with samples generated one at a time,
ran the first 111 questions that way, then lowered the limit to 4{,}096; all
111 finished well within either limit. At the
longer limit the unedited model scores 25.6\% (pass@4 44.9) and the edit
25.4\% (pass@4 47.5), differences of $-0.3$ ($-3.5$, $+3.0$) and $+2.5$
($-3.5$, $+8.6$) points, and 2.9\% and 2.7\% of samples are unfinished.
Doubling the limit changes neither model's score by more than a point.

On GSM8K the prompt is the question followed by ``Please reason step by step, and
put your final answer within \verb|\boxed{}|.'', in Qwen3's chat template with
thinking turned off, sampled at temperature 0.7, top-$p$ 0.8 and top-$k$ 20 for
up to 512 new tokens. The answer is the last \verb|\boxed{}| expression, or the
last number if there is none. The edit is applied to every attention and MLP
projection in fp32 and the model then runs in bf16. Rounding then distorts the
edit: the error $\|\mathrm{bf16}(W') - W'\|_F$, where $W'$ is the edited
matrix, is on average 25\% (16\% at $\lambda=0.2$) of the intended change
$\|W' - W\|_F$ for Timber's
tail edit, 4\% for the stable-rank tail edit, 5.5\% for uniform scaling
matched to the stable-rank edit and 88\% for uniform scaling matched to
Timber's edit, which bf16 storage almost erases. The uniform scaling factor is
computed per matrix; its mean is 0.75 for the stable-rank match. The pass@$k$
values use the unbiased estimator of \citet{chen2021codex} with $n=8$.

\begin{table}[H]
\caption{All edit variants on the first 300 GSM8K test problems, pass@$k$ in
percent. Differences
from the unedited model in points, with 95\% paired bootstrap intervals,
not corrected for the fourteen comparisons. ``Answered'' is the share of
samples with a final boxed answer. ``Uniform'' scales the whole update, per
matrix, to the size of the named tail edit.}
\label{tab:cutoff300}
\begin{center}
\small
\begin{tabular}{lccccc}
\toprule
edit & pass@1 & pass@8 & answered & $\Delta$ pass@1 & $\Delta$ pass@8 \\
\midrule
none                                  & 63.4 & 84.7 & 96.5\% & -- & -- \\
tail $\times 0$, eRank cutoff         & 63.9 & 87.7 & 96.7\% & $+0.5$ ($-1.3$, $+2.3$) & $+3.0$ ($-0.3$, $+6.3$) \\
tail $\times 0.2$, eRank cutoff       & 64.0 & 85.7 & 96.8\% & $+0.5$ ($-1.3$, $+2.4$) & $+1.0$ ($-2.0$, $+4.0$) \\
tail $\times 0.5$, eRank cutoff       & 64.7 & 88.0 & 96.8\% & $+1.3$ ($-0.5$, $+3.0$) & $+3.3$ ($+0.7$, $+6.3$) \\
tail $\times 0$, stable-rank cutoff   & 59.3 & 83.0 & 90.6\% & $-4.2$ ($-6.9$, $-1.4$) & $-1.7$ ($-5.3$, $+2.0$) \\
tail $\times 0.5$, stable-rank cutoff & 64.5 & 87.3 & 95.1\% & $+1.1$ ($-1.1$, $+3.3$) & $+2.7$ ($-1.0$, $+6.3$) \\
uniform, matched to eRank edit         & 64.2 & 86.3 & 96.5\% & $+0.8$ ($-0.8$, $+2.4$) & $+1.7$ ($-1.0$, $+4.3$) \\
uniform, matched to stable-rank edit   & 66.0 & 88.7 & 96.0\% & $+2.6$ ($+0.4$, $+4.8$) & $+4.0$ ($+1.0$, $+7.0$) \\
\bottomrule
\end{tabular}
\end{center}
\end{table}